\documentclass[10pt, a4paper, twocolumn]{article} 
\usepackage[style=apa, backend=biber]{biblatex}
\usepackage{color, soul}       
\usepackage[british]{babel} 
\usepackage{algorithm}
\usepackage{algorithmic}
\usepackage{csquotes}
\usepackage[margin=0.75in]{geometry}

\usepackage{graphics} 
\usepackage{epsfig} 
\usepackage{mathptmx} 
\usepackage{times} 
\usepackage{multirow}
\usepackage{amsmath}
\usepackage{amssymb}  
\usepackage{orcidlink} 
\usepackage{authblk} 
\usepackage[font=footnotesize]{caption}

\let\cite\parencite

\title{
PathSpace: Rapid continuous map approximation for efficient SLAM using B-Splines in constrained environments
}

\author[1]{Aduen Benjumea}
\author[1]{Andrew Bradley}
\author[1]{Alexander Rast}
\author[1]{Matthias Rolf}
\affil[1]{School of Engineering, Computing \& Mathematics, Oxford Brookes University, Oxford, UK}

\date{}

\begin{document}

\maketitle

\begin{abstract}

Simultaneous Localization and Mapping (SLAM) plays a crucial role in enabling autonomous vehicles to navigate previously unknown environments.
 Semantic SLAM extends visual SLAM, leveraging the higher density information available to reason
about the environment in an action-oriented manner. This allows for better decision making by exploiting prior structural knowledge of the environment, usually in the form of labels. 
Current semantic SLAM techniques mostly rely on a dense geometric representation of the environment, limiting their ability to apply constraints based on context.
We propose PathSpace, a novel semantic SLAM framework that uses continuous B-splines to represent the environment in a compact manner, while also maintaining and reasoning through the continuous probability density functions required for probabilistic reasoning.
This system applies the multiple strengths of B-splines in the context of SLAM to interpolate and fit otherwise discrete sparse environments. We test this framework in the context of autonomous racing, where we exploit pre-specified track characteristics to produce significantly reduced representations
at comparable levels of accuracy to traditional landmark based methods and demonstrate its potential in limiting the resources used by a system with minimal accuracy loss.

\end{abstract}

\section{INTRODUCTION}

Autonomous agents employ Simultaneous Localisation and Mapping (SLAM) as the probabilistic framework to reason through sensor and actuation uncertainty, 
producing a reliable belief of an agent's state and its environment in previously unknown spaces and aid in later navigation. While specific implementations vary across platforms, 
from small indoor robots to large autonomous vehicles, they share a core reliance on these principles for navigation. 
Visual SLAM (VSLAM) often utilizes dense geometric data from visual sensors to optimize relative positions \cite{ABASPURKAZEROUNI2022117734}. 
However, these representations frequently fail to capture the contextual information essential for high-level cognition. 
Semantic SLAM bridges this gap by incorporating object labelling and segmentation directly into the localisation process \cite{rs14133010}. 
By moving beyond purely geometric data, the system becomes more robust to granular visual shifts like lighting or perspective changes, 
allowing the agent to interpret its surroundings conceptually and facilitate more complex, reactively informed tasks.

Semantic SLAM in principle works by leveraging known context about the environment, which has the downside of being more restrictive to a given domain 
for the benefit of adding valuable contextual information to the system \cite{s21196355}. Generally speaking, the more tightly structured the domain, the more context can be leveraged 
as more assumptions of the environment can be made in advance. The idea of anchoring \cite{coradeschi2003introduction} applied to an indoor domain, for example, 
can relate object detections to label an entire room at a higher abstracted level, usually with the goal of improving human to robot 
interfaces \cite{galindo2005multi}. Similar reasoning can be applied to autonomous vehicles, where the known structure of the road can greatly limit 
the domain of plausible layouts or be used for labelling entire areas of the environment as requiring a higher level of attention \cite{roth2024viplanner}. In the context of roads, 
these can be areas of likely higher traffic, or areas with rapid changes in curvature in the case of high performance scenarios.

\begin{figure}[tp]
        \centering
        \includegraphics[trim= 0 0 0 50, clip, width=\linewidth]{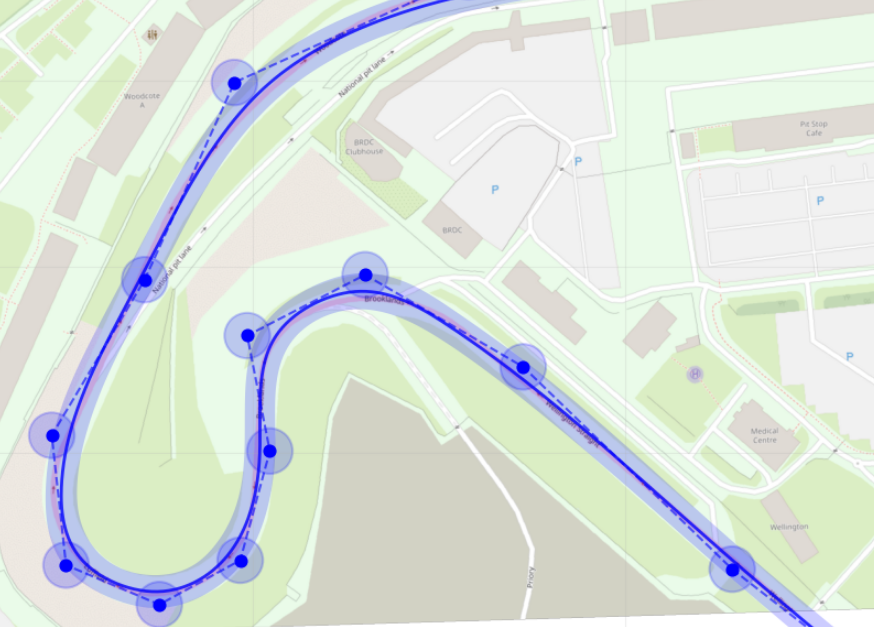}
        \caption{Example of a continuous B-spline representation of using a limited number of control points for its representation. Base map data by OpenStreetMap (available under the Open Database License)}
        \label{header_pic}
\end{figure}

B-splines are a staple of geometric representation in CAD design and computer graphics with strengths including, but not limited to, their smoothing, interpolating and extrapolating capabilities.
Despite these advantages, they are not exploited to their full potential in SLAM systems.
The few examples seen as part of a SLAM system often condense rich point cloud data of the environment as several curve objects \cite{Pedraza_2009}. 
Their use as the main representational method of an environment defined by sparse or semantic 
features has, however, not been explored. Cubic B-splines possess C2 continuity, guaranteeing a smooth second derivative, which in turn translates to an equivalent 
smooth acceleration curve and therefore preventing the representation of physically impossible motion paths \cite{9782519}. Additionally, the local control property of 
B-splines means an update to a localised area of the curve does not need to affect the entire curve, making it more suitable for live updates given previously seen samples. 
A representation of, for example, a path, track, or otherwise traversable area using a single spline would, by design, reduce the degree of freedom of landmark association and allow for the representation of 
seamlessly closed circuits. As we can see in Fig. \ref{header_pic}, in the world of motorsport, this can lead to maps being represented much closer to how lap time and race line optimizations operates, making the map of the system inherently contain semantic information for the task at hand
and therefore facilitate integration of optimal paths as a way to allow higher level cognition in the system. Ultimately, such representations can more closely model how a human would understand
the environment, and a system with some attention to curvature would inherently allow for embedding maximum speed controls as an example, providing more context for path planning and control sub-systems.


The main contributions of this paper can be summarized as follows:
\begin{itemize}
    \item A novel framework for the use of B-splines to represent continuous probability distributions functions from discrete landmarks in online SLAM systems.
    \item A spatially aware optimization system that enforces a representational budget, limiting computational growth based on environmental complexity rather than raw landmark density.
    \item An online feasibility study comparing the proposed system against a traditional method both in performance and scalability within the context of high performance autonomous vehicles.
\end{itemize}


\section{RELATED WORK}
\label{sec:related_work}
\subsection{SLAM in Autonomous Vehicles}
An autonomous vehicle (AV), much like any autonomous mobile robot tasked with navigation, 
requires four steps to complete its task: perception, localization, cognition and control \cite{siegwart2011introduction}. Localization here is a subtask of SLAM, 
which leads to the recent interest in advancing these technologies alongside the increased interest in commercial AVs.

A SLAM system can be thought of as offline, where the prediction over the entire history of positions is optimized, or online, 
where we treat localization as a Markovian process and optimize the current position \cite{thrun2005probabilistic}. 
Generally, live systems concerned with autonomy and reactiveness will implement online methods with higher updates rates and greater constraints
when it comes to computational resources available.

We can categorize SLAM methods by their mathematical basis. Filter based systems such as the extended Kalman filter (EKF) \cite{thrun2002probabilistic} 
solve SLAM recursively, processing the state of the vehicle and the map as a prediction or prior and an update based on sensor readings. 
Variants such as the unscented Kalman filter (UKF) \cite{wan2000unscented} provide alternative methods for the linearization of the models involved for 
increased robustness. Highly non-linear environments frequently opt for a particle filter (PF) approach instead, where smaller state predictions are made over many 
hypotheses referred to as particles, such as FastSLAM \cite{montemerlo2003fastslam}. 
In optimization based approaches, such as Graph SLAM \cite{grisetti2011tutorial}, 
a front end collects all the system constraints and a backend performs non-linear optimization to obtain a maximum likelihood of states.

\subsection{Semantic SLAM}

The emergence of high-speed, specialised visual processing in GPUs has resulted in standardised VSLAM approaches, 
often leveraging bundle adjustment\cite{triggs1999bundle} to perform visual odometry \cite{nister2004visual}, 
such as the well known ORB SLAM \cite{mur2015orb}. Similar techniques can be applied to other sensors such as LiDAR \cite{shan2020lio} or depth cameras \cite{yu2018ds}.

Furthermore, deep learning has enabled the introduction of object detection and segmentation to these tasks, 
leading to the field of semantic SLAM extending VSLAM \cite{rs14133010}. 
Providing higher level information into the system decreases the dependence on geometric points,
and therefore subtle changes in the sensors domain generally have a lesser impact on the system integrity \cite{you2022misd}. 
Systems using semantic SLAM range from discrete landmark based systems \cite{bowman2017probabilistic}, 
precise object characteristic recognition \cite{pavlakos20176} to continuous or finely detailed segmentation of the environment \cite{Tateno_2017_CVPR}. 
Semantic SLAM is currently at the frontier of SLAM research \cite{rs14133010}, and more complex methods aim at 
facilitating more complex tasks and increasing autonomy by more closely relating it to the path planning and decision 
making tasks or allowing for more flexibility in dynamic environments \cite{hu2024semantic}.

While VSLAM brings the steps of perception and localization closer together, semantic SLAM can be thought of as bringing the steps of localization and cognition closer together,
as it allows for the system to leverage the context of the environment in its localization process, and therefore allows for the system to make more informed decisions based on this context.

\subsection{B-splines in SLAM}
B-splines are commonly used in product design for their ability to produce continuous surfaces, for 
light rendering or in animation to produce smooth motion speeds \cite{hasan2024b} and exploit their local control and flexible level of continuity for interpolation with limited parameters. In SLAM however, 
they are more often used as a means to represent large and dense clusters of point reading on each update cycle as a single condensed object \cite{Rodrigues_2021}\cite{liu2009new}\cite{Kanna_2024}\cite{pedraza2007bs}. 
This reduces the number of redundant points being stored and ultimately reduces unwanted noise by smoothly 
fitting the data to a simpler surface. When it comes to motion, B-splines have been used in a similar fashion 
to reduce unwanted prediction noise for the predicted path of an agent \cite{liu2010towards}, or to interpolate vehicle motion predictions for the purposes
of handling asynchronous measurements or issues such as rolling shutter measurements \cite{lovegrove2013spline}.
Another use of B-splines is as surfaces, representing what would otherwise be a discrete occupancy grid map as a continuous surface, 
which can be leveraged for better predictions in path planning and obstacle avoidance \cite{Kanna_2024}.

While SLAM research recognizes B-splines for their smoothing and compression properties, particularly in representing dense point clusters, their application within semantic frameworks has not yet been fully exploited, nor have online probabilistic frameworks for them been developed. 
Specifically, a gap exists in creating more adaptable systems able to react to the overall semantic properties of the environment while discerning a continuous probabilistic understanding of the environment solely from discrete features.
By utilizing the representational flexibility of B-splines, we can develop a task-oriented probabilistic system that prioritizes environmental features most relevant to the agent, focussing on an alternate representation of the environment that
 are most suited to organize driving behaviour through path planning and control.

\section{B-SPLINE REPRESENTATION}
\label{sec:b_spline_representation}
\subsection{Fundamentals}
A B-spline or basis spline is a piecewise polynomial function  $\mathbf{S} = \{k, \mathbf{t}, \mathbf{C}\}$ defined by an integer order $k$ (degree $k-1$), 
a knot vector $\mathbf{t} = [t_0, t_1, \ldots, t_n]$ and a set of control points (or coefficients) $\mathbf{C} = [c_0, c_1, \ldots, c_n]$ 
where each $c_i$ has the same dimensionality as the output of the spline. The B-spline at input parameter value $u$ is 
defined as:
\begin{equation}
    S_{\mathbf{t},k}(u) = \sum_{j=0}^{n} C_j \beta_{j,k}(u) \label{eq:b_spline}
\end{equation}

Here, $\beta_{j,k}(u)$ are the B-spline basis functions defined by the 
Cox-de Boor recursion \cite{piegl2012nurbs} as follows for $k = 0$ and $k > 0$ respectively:

\begin{equation}
    \beta_{j,0}(u) = \begin{cases} 1 & \text{if } t_j \le u < t_{j+1} \\ 0 & \text{otherwise} \end{cases} \label{eq:b_spline_basis_0}
\end{equation}
\begin{equation}
    \beta_{j,k}(u) = \frac{u - t_j}{t_{j+k} - t_j} \beta_{j,k-1}(u) + \frac{t_{j+k+1} - u}{t_{j+k+1} - t_{j+1}} \beta_{j+1,k-1}(u) \label{eq:b_spline_basis_k}
\end{equation}

This can be thought of as a smooth average weighted by $\beta$ of $k$ control points $C_i$ nearest to $u$. 
The knot vector $\mathbf{t}$ effectively tells us the range influence each control point has on the distance 
from $u$ to the control points, and does not need to be uniform nor requires values to be unique, 
but it must be non decreasing ($t_i \le t_{i+1}$). The multiplicity of each value effectively reduces 
the degree of the spline at that particular point, allowing us to represent discontinuities within the spline if needed. 
If the multiplicity of values at the ends of the knot vector equals the order $k$, the spline is said 
to be clamped (or left/right clamped if asymmetrical) \cite{piegl2012nurbs}, and it guarantees the ends of the spline will 
interpolate the end control points. We assume a spline will be clamped unless otherwise stated. The processes 
of clamping or unclamping, then, refer to the increase or decrease of the multiplicity to 0 or the order of the spline of the end knot values respectively. 
This would often be paired with a recalculation of the control points to preserve the shape of the spline.

\subsection{Representing uncertainty}
To fully benefit from the the continuous representation B-splines provide for the mean prediction of the environment, 
we need to represent the uncertainty in an equally continuous manner, allowing any sample along the spline to be associated with a probability distribution.

In line with the rationale of Kalman Filter methods, we represent uncertainty as a single covariance matrix $\varSigma$.
This matrix tracks the marginal uncertainties of both the agent state and the individual control points, while the off-diagonal cross-covariance terms preserve the dependencies necessary for optimization.

\begin{figure}[tp]
        \centering
        \includegraphics[width=\linewidth]{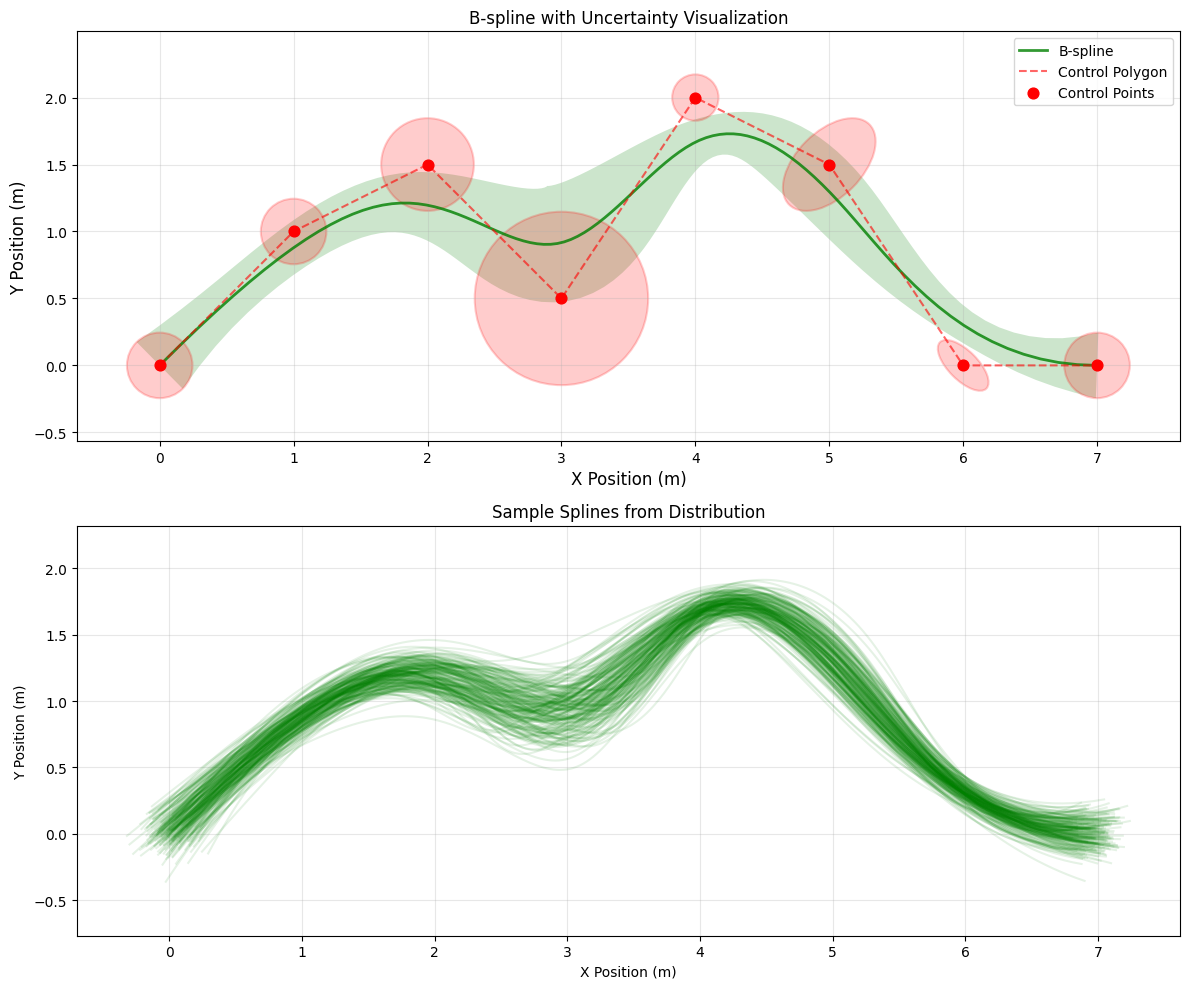}
        \caption{Demonstration of continuous uncertainty representation from discrete control point uncertainties and how it creates a continuous distribution of splines.}
        \label{uncertaint_demo}
\end{figure}

Since each control point provides direct local control of the spline, we associate a Gaussian distribution to each control point, and propagate this uncertainty through the same basis functions that govern the spline itself.
Thus, we determine the uncertainty of a point on the spline $\varSigma_{S(u)}$ given the uncertainty of the control points by squaring the basis functions as follows:
\begin{equation}
    \varSigma_{S(u)} = \sum_{j=0}^{n} \varSigma_{C_j} (\beta_{j,k}(u))^2
\end{equation}

where $\varSigma_{C_j}$ is the marginalised covariance matrix of the control point $C_j$ within the larger $\varSigma$ covariance matrix of the entire system, 
treating the spline as a continuous uncertainty distributions as is illustrated in Fig. \ref{uncertaint_demo}.

\subsection{Expansion}

\begin{figure}[tp]
        \centering
        \includegraphics[width=\linewidth]{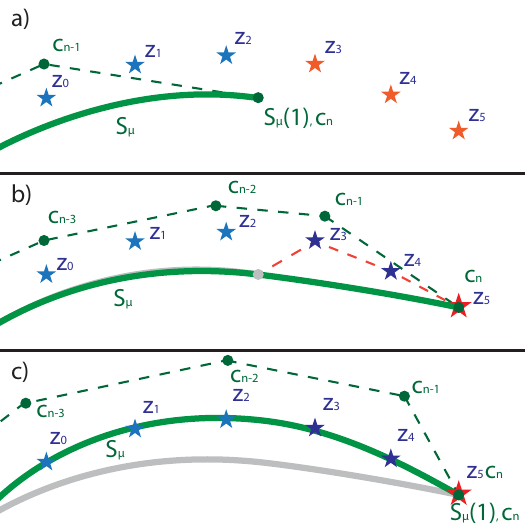}
        \caption{Process of expanding the spline $S_z$ based on new readings $\mathbf{z}$. Each reading $\mathbf{z}_i$ are categorised as expansion points of the closest spline point is $\mathbf{S}(1)$ and is beyond a threshold distance from $\mathbf{C}_n$ and $\mathbf{C}_{n-1}$ or as update points otherwise (a), 
        the interpolation point is selected and the rest are re-categorised while the spline extends to interpolate the new point (b), 
        and an update is performed with all remaining points (c).}
        \label{expansion_process}
\end{figure}

Upon the detection of a set of landmarks, the system determines if an extension of the spline is required and where to extend it.
Fig. \ref{expansion_process} illustrates the steps taken to determine and handle an extension of the spline, which begins by determining when a landmark is updating or extending the spline.
In order to perform this classification, we employ a function where a measurement point is 
marked for expansion of the spline when it satisfies a three-part boundary condition: the projection of the landmark onto the spline, i.e. 
the closest point in the spline must coincide with the terminal parameter $u=1$, the Euclidean distance to this endpoint must exceed a predefined growth 
threshold, and the point must maintain sufficient separation from the second-to-last control point $C_{n-1}$ to prevent localized over-parameterization 
while also avoiding over-stretching of the particular spline segment. This heuristic can be parameterized as informed by prior knowledge and assumptions of the task domain. There is also potential for more complex functions if more constraints can be gleaned from the environment.

Before the update is performed, the points marked for expansion are ordered by a proximity chain, starting from the reading closest to $\mathbf{S}(1)$ followed by the closest to the last point in the sequence repeatedly.
The last point in this sequence is used to extend the spline, while the rest are re-categorised as update points.

We extend the spline to this point following the methodology described in \cite{hu2002extension}. This process involves unclamping the spline by reducing the 
multiplicity of the terminal knot value, followed by a recalculation of the local control points to preserve the original shape. The extension point is then 
inserted as a control point along with uncertainty associated with the sensor reading, and the knot vector $\mathbf{t}$ is re-clamped around it to ensure exact interpolation at the 
new boundary. Once added, the spline can be updated to fit the remaining points as shown in Fig. \ref{expansion_process}

\subsection{Update}
When one or more landmarks $\mathbf{z}$ are detected and determined to belong to the existing spline, we initially fit a continuous spline with the discrete readings received. 
For this process, concrete data association is not required, as we are interested in what the landmark is 
communicating about the environment rather than its specific identity. For this same reason, this transform will combine multiple landmark readings into 
a singular, continuous spline reading, and thus only a single update will be performed to the relevant control points per perception cycle. 

\begin{figure}
    \centering
    \includegraphics[width=\linewidth]{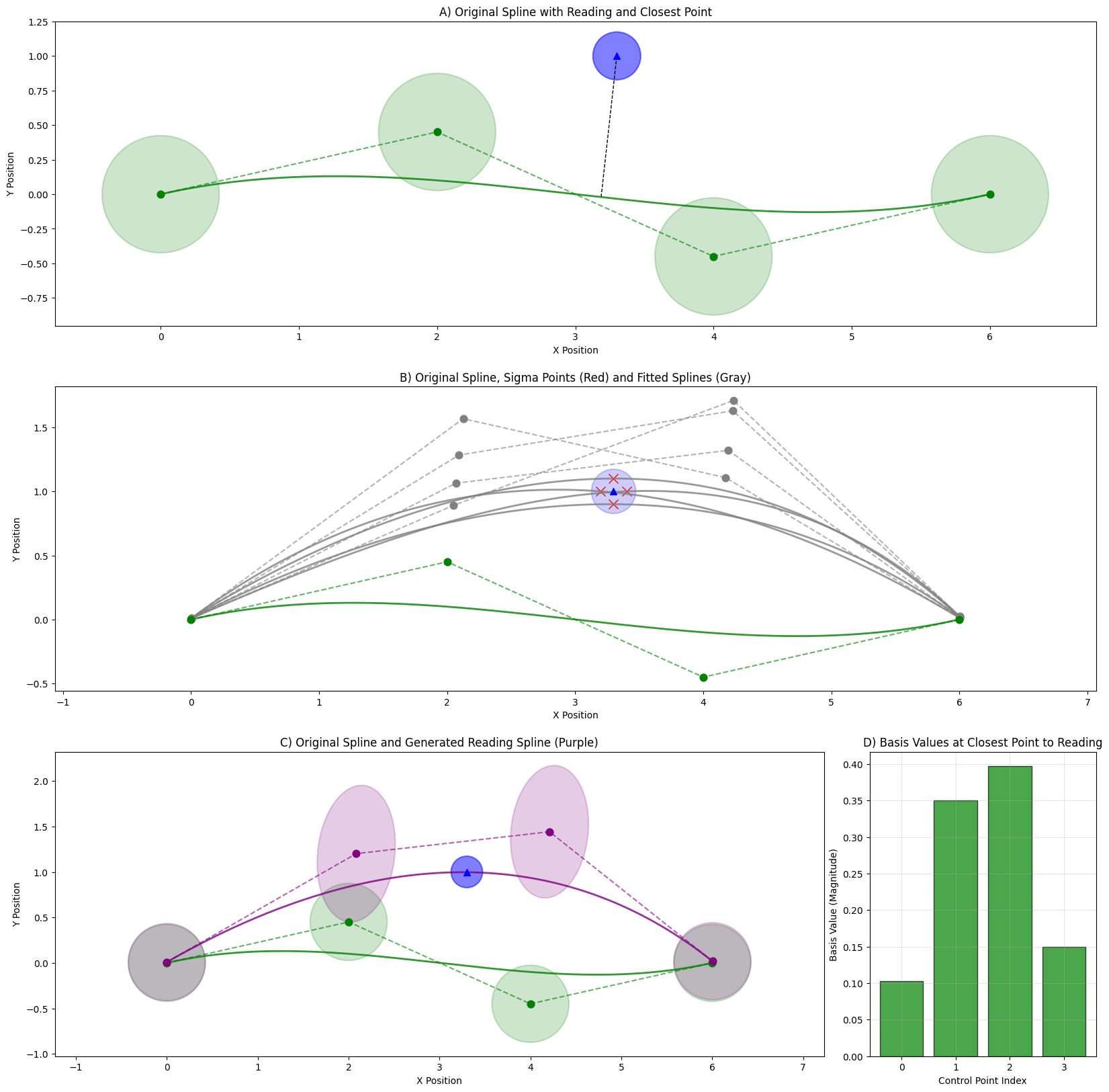}
    \caption{Demonstration of $\mathbf{S}_z$ fitting. In (A) we find the closest point in $\mathbf{S}_\mu$ (green) to a new set of readings $\mathbf{z}$ (blue). 
    In (B) we generate the sigma points $\mathbf{X}$ (red) based on the uncertainty of the reading and fit $\mathbf{S}_\mu$ to each one using ridge regression fitting to generate the $\mathbf{Z}$ splines (gray).
    in (C), the the control points in $\mathbf{Z}$ are averaged to obtain the control points and their respective covariance, defining $\mathbf{S}_z$ (purple).
    (D) shows the basis values of each control point on the closes spline point to the reading, determining how much each control point will be allowed to move to fit in (B)}
    \label{fig:z_spl}
\end{figure}

Firstly, we create a copy of the current state spline $S_\mu$. This process will mutate this state to agree with the perceived data. We refer to this newly fit spline as $S_z$. Fig. \ref{fig:z_spl} shows this process.
We locate the parameter values $u_z$ along $S_\mu$ that would generate the closest point to the reading on the spline, $\mathbf{\bar{z}}$, and note the basis values for each control point used at those 
$u$ values by our spline as $\mathbf{b}_z$. We use this to populate each row basis matrix $\mathbf{B}$. This matrix is of size $m \times n$ where $m$ is the number of readings and $n$ the number of control points.

\begin{equation}
    \min_{\mathbf{C}} \|\mathbf{y} - \mathbf{B} \mathbf{C}\|^2 + \|\mathbf{L}(\mathbf{C} - \mathbf{C}_\mu)\|^2
\end{equation}

Where $\mathbf{C}_\mu$ is the current position of the control points in the state spline. $\mathbf{L}$ in this case is the regularization parameter, 
which can often be found expressed as $\mathbf{I}\lambda$ when a uniform regularization is sufficient. In our case however, we require the control points with a higher influence over 
$\mathbf{\bar{z}}$ to be regularized less, meaning they will have a bias to travel further when fitting. To achieve this, we define it as 

$$\mathbf{L} = \lambda \operatorname{diag} \left( \mathbf{1}_n - \frac{1}{m} \mathbf{B}^T \mathbf{1}_m \right)$$

with $\mathbf{1}_m$ and $\mathbf{1}_n$ as column vectors of 1s of the respective size and $\lambda$ still as a regularization scaling parameter. We then solve for the control points of the fitted spline as

\begin{equation}
    \mathbf{C} = (\mathbf{B}^T \mathbf{B} + \mathbf{L})^{-1} (\mathbf{B}^T \mathbf{y} + \mathbf{L} \mathbf{C}_\mu) \label{eq:ridge_regression}
\end{equation}

This method does not aim to fit the readings exactly, but rather generates a constrained line of best fit. 
This has the additional effect of smoothing over contradicting or noisy reading points in benefit of a smooth semantic understanding of the traversable area.

In order to maintain the uncertainty of the reading in this new representation and be able to later use it in the update step of a Kalman filter, we need to be able to propagate the uncertainty of the reading through this fitting process.
We do this through the use a Cubature Transform \cite{4982682}, which is a method for propagating uncertainty through non-linear transformations, similar to the Unscented Transform used in the UKF but with a different sampling strategy that can provide better performance in higher dimensions.
This process involves the generation of a set of sigma points $\mathbf{X}$ that are representative of the distribution of the reading, which are then passed through Eq. \ref{eq:ridge_regression}, defined as the aforementioned ridge regression process, to obtain a new set of points in the spline representation $\mathbf{Z}$. The mean and covariance
after this transformation can then be calculated as a weighted average of the transformed points, with weights determined by $\mathbf{X}$. More detail on the formulation of these processes can be found in \cite{wan2000unscented}

From this point, the control point values and their uncertainty can be used directly in the update step of a Kalman filter.
The control point association between the reading and state are direct and do not require additional processing.
Traditional discrete methods require some form of matching optimization for landmark association when the landmarks are not uniquely identifiable.
This process, by contrast, uses whichever point along the continuous spline is closest to the discrete reading. This means the fitted spline is able to capture how the uncertainty on a landmark interacts with the current direction of the spline,
as the sample based transform realises less influential uncertainty directions as shown in Fig. \ref{fig:z_spl}. This also directly tackles difficulties with complex data association and avoids duplicate landmarks in the map from misassociation.

\subsection{Simplification}
\begin{figure}
    \centering
    \includegraphics[width=\linewidth]{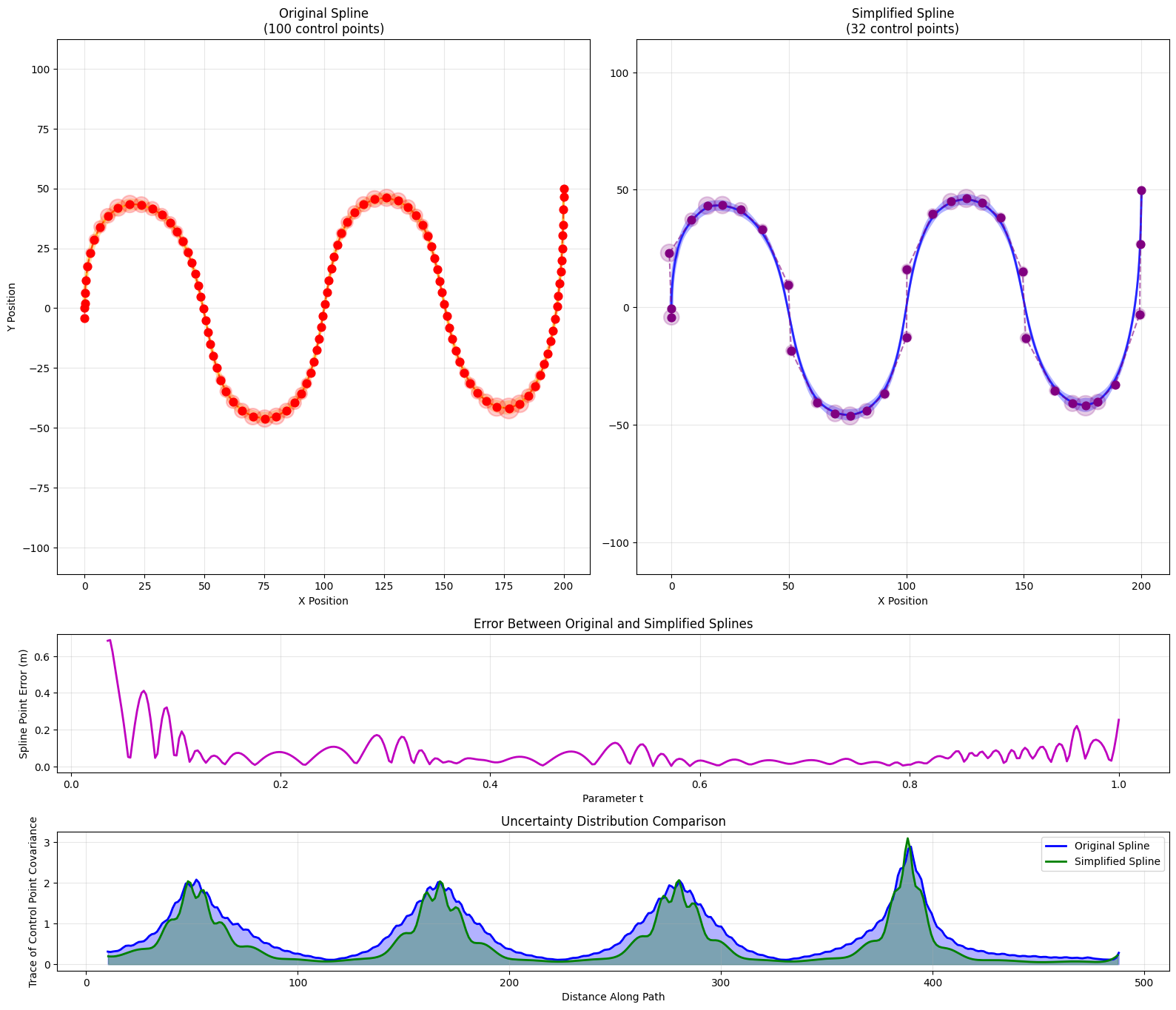}
    \caption{Comparison between sample spline before and after simplification. We show a control point reduction from 100 to 32 points with minimal data loss}
    \label{fig:simplification}
\end{figure}
To maintain map sparsity and reduce memory consumption, we employ a curvature weighted simplification that re-allocates representational 
capacity based on geometric complexity. The process begins by uniformly sampling the existing spline to derive a localized curvature profile, 
which is then normalized into a probability distribution. A linear baseline is integrated into this distribution to ensure a minimum control point density 
remains in low-curvature or linear segments, preventing topological gaps. By distributing a target number of knots (determined by the length and complexity of the features of the operational domain) across the resulting cumulative curvature 
distribution, the algorithm effectively concentrates control points in complex regions such as sharp corners while simplifying redundant data in straight sections. 
Finally, a new parametric model is generated through a least-squares optimization that fits the reduced set of control points to the original sampled data, ensuring the 
global shape is preserved while significantly lowering the dimensionality of the map state. Similarly to how we propagate error through our $\mathbf{S}_z$ fitting, 
we propagate uncertainty of the overparameterized spline through this process using a sample based Cubature Transform, generating sigma points for the state of the spline, 
performing the simplification once the ideal knot vector has been determined and recollecting the mean and covariance of the simplified spline. This process is performed at a low
frequency as it optimizes the entire spline and incurs a considerable computational overhead, but provides a near equivalent representation of the map so far with significantly fewer parameters. Fig. \ref{fig:simplification} compares a spline before and after this operation is performed.

\subsection{Experiments}
\subsubsection{Isolated tests}
To validate the core operations of this framework, we experimentally test each of the components described in the previous sections in isolation, 
using synthetic data to validate the expected behaviour of each of these elements and obtain an initial understanding of useful parameterization and/or possible implementation considerations.
\subsubsection{Autonomous Racing}
As a proof of concept, we apply this framework to the context of autonomous racing, where the track can be represented as a single closed loop spline.
The application in the context of Formula Student Driverless a highly controlled environment with circuits demarcated by brightly colored traffic cones, but where the visual features of the track are not guaranteed to be consistent across laps,
marked as brightly colored cones. In collaboration with the Oxford Brookes Racing Autonomous Team, we build on top of a pre-existing front end, providing 3D cone detections semantically with their respective label and uncertainty measure.
The goal in this scenario is to map the differently coloured cone boundaries (blue and yellow), demonstrating feasibility, stability and resource usage of this methodology when applied to a real-world application requiring the use of a computationally-efficient SLAM approach.

\section{Results}
\label{sec:results}
\subsection{Simulation setup}
\begin{figure}[tp]
        \centering
        \includegraphics[width=\linewidth]{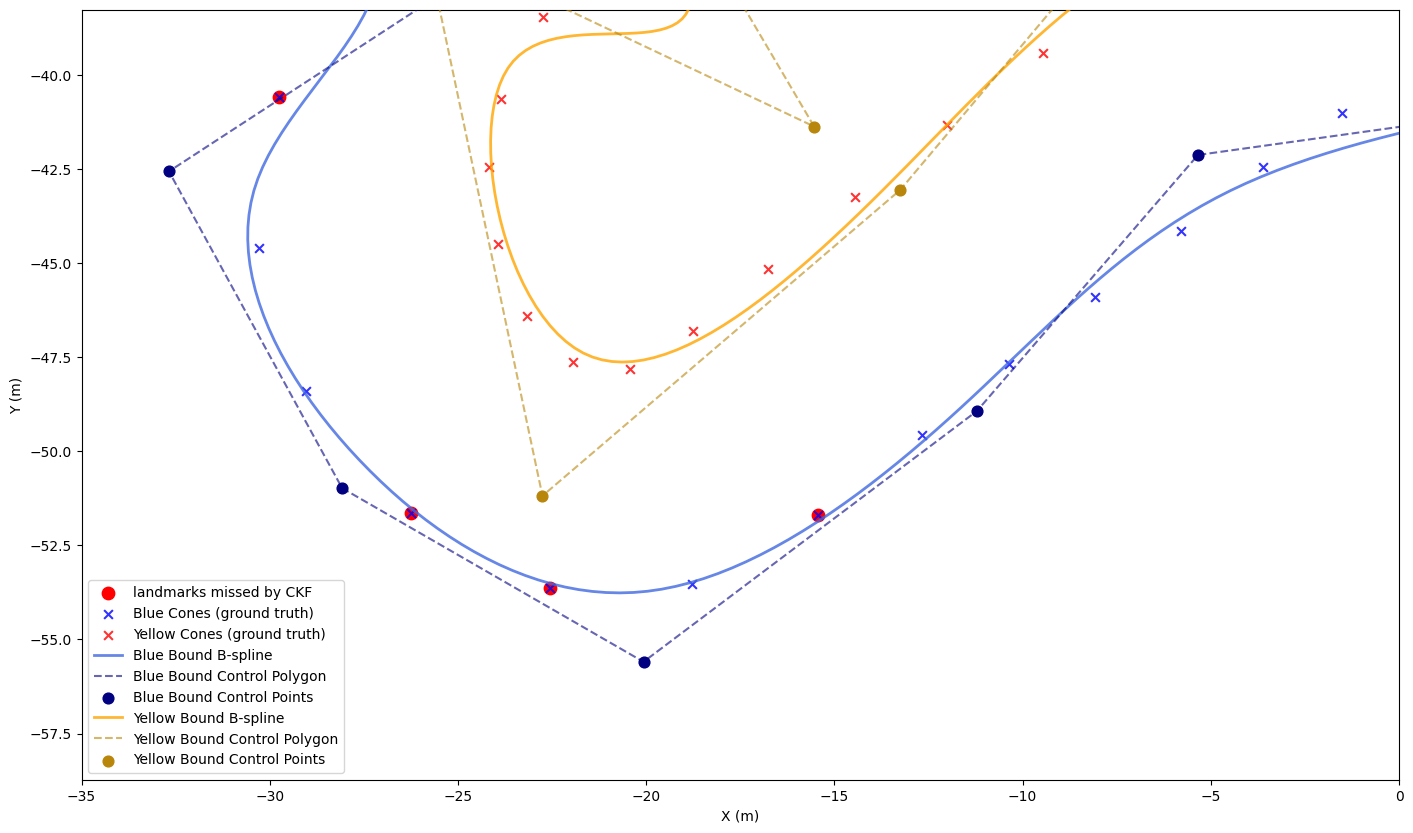}
        \caption{Section of the mapped simulation track, highlighting the PathSpace representations of both boundaries and examples of where CKF misses a landmark while PathSpace interpolates}
        \label{fig:sim_setup}
\end{figure}

Following the case study of autonomous racing, we generate a synthetic track with similar characteristics to those found in the real application based on technical specification of the competition and reconstructions from aerial images taken in real world testing. This includes the separation between landmarks (cones), track boundaries and overall track length,
as well as the expected track elements such as straights, hairpins and chicanes. For the purpose of testing the core functionality of the system, we isolate the backend by assuming a known front end, providing cone detections with their associated 
uncertainty as well as providing odometry data for the agent. We then allow the agent to run several laps around the track, noting the scale and accuracy of the map generated. Fig. \ref{fig:sim_setup} shows this layout.

As a baseline, we compare our B-spline system against a traditional landmark-based representation of the track, implementing a CKF with a similarly structured state and overall implementation style. 
Since landmarks in this case study are not uniquely identifiable, we perform data association in this baseline using the Hungarian algorithm. This solves a set matching optimization
given a cost matrix. In this case, we provide this cost as the Mahalanobis distance between mapped landmarks and perceived landmarks to account for evolving landmark certainty
and more accurately model landmark association as it would be implemented in a live system. 

\subsection{Simulation results}

We run our systems on the same simulated environment with the same given odometry and front end readings and compare their behaviour and performance in table \ref{tab:results}. 
Error is calculated as RMSE to the ground truth across all landmarks while keeping track of landmarks that have not been accounted for by each system if there are no mapped points within a 3m threshold. the reference points in the case of PathSpace is the closest point along the spline to a given ground truth landmark.
The size of the map is represented as number of control points in the case of PathSpace, or number of landmarks in the case of CKF. As they both represent discrete geometrical points and occupy the same dimensions, we compare them equivalently. 

Once PathSpace detects the completion of the first lap by the path length exceeding a threshold and an association with an early segment of the spline is made, it closes the spline into a loop and performs a simplification cycle. After this point, no new landmarks are added, as the spline cannot be extended.
We see this in the number of landmarks remaining at the optimal size determined by this system beyond the first lap, while CKF continues to expand its map. PathSpace successfully represents all landmarks while only requiring 44\% of the parameter space as our baseline. CKF in contrast misses a considerable number of landmarks due to blind spots, even after several laps.
The interpolation performed by PathSpace allowing it to map beyond range of sight can be considered a form of semantic embedding, as we are able to estimate the shape of the space given we know of its continuity as a prior.

CKF is also susceptible to ghosting, where a landmark is mapped more than once due to a missasociation and requires an additional process to be removed from the map. 
PathSpace continuous nature makes it resistant to this effect, as not discrete association is performed.


PathSpace achieves an average of 16.8\% improvement in RMSE over CKF. The smoothing and interpolation performed by PathSpace allows it to more closely approach the ground truth of any given point, effectively reducing one degree of error. 
Recognizing that professional racing lines typically vary by more than $\pm 1m$ as supported by \cite{brayshaw2005quasi}, the accuracy of both PathSpace and our baseline remains well within the functional tolerances for this case study. 

\begin{table}[tp]
\centering
\caption{Comparison of accuracy, map size, and landmark coverage: PathSpace B-spline system vs. baseline CKF on a 200 landmark circuit.}
\label{tab:results}
\resizebox{\columnwidth}{!}{%
\begin{tabular}{|c|c|c|c|c|c|c|}
\hline
\textbf{Lap} & \multicolumn{3}{c|}{\textbf{PathSpace}} & \multicolumn{3}{c|}{\textbf{CKF}} \\
\cline{2-7}
 & \textbf{RMSE (m)} & \textbf{Size} & \textbf{Missed} & \textbf{RMSE (m)} & \textbf{Size} & \textbf{Missed} \\
\hline
1 & 0.582 & 126 & 0\% & 0.729 & 158 & 21\% \\
\hline
2 & 0.624 & 76 & 0\% & 0.722 & 166 & 17\% \\
\hline
3 & 0.611 & 76 & 0\% & 0.720 & 169 & 15.5\% \\
\hline
4 & 0.589 & 76 & 0\% & 0.710 & 171 & 15\% \\
\hline
5 & 0.574 & 76 & 0\% & 0.710 & 173 & 14\% \\
\hline
\end{tabular}
}
\end{table}

\subsection{Scalability}

\begin{figure}[tp]
        \centering
        \includegraphics[width=\linewidth]{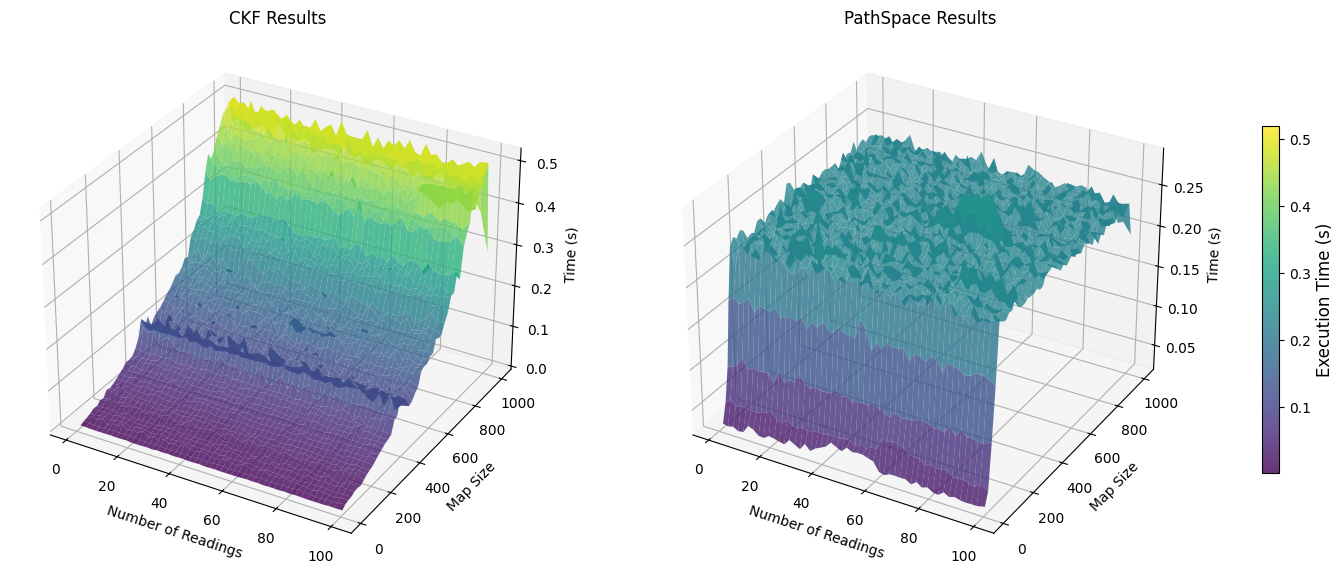}
        \caption{Scalability comparison of a traditional CKF implementation (top) against our B-spline representation (bottom) in terms of both number of landmarks in the map and number of landmarks observed per update. each point represents the average execution time of 10 update cycles.}
        \label{figurelabel5}
\end{figure}

We further investigate the scalability of our B-spline system against a traditional point-based CKF implementation in terms of both the number of landmarks in the map and the number of landmarks observed per update.
in Fig. \ref{figurelabel5}, we compare the execution time for an update cycle in similar conditions for a wide range of parameter combinations.
These results demonstrate that while PathSpace has a higher initial cost due to the fitting and transformation into the spline representation, this same representation
limits the computational cost of the update step as the number of landmarks increases based on the expansion and simplification processes, while the point-based method scales cubically. This is due to PathSpace's ability to simplify the map and interpolate between landmarks.
This essentially sets a maximum complexity budget for the system, and therefore the execution time flattens when the system finds an optimal representation for the given layout.
It is worth noting that the specific complexity at which PathSpace plateaus will depend on the parameterization of the system and can therefore be optimised further for a specific system. 
The number of readings per update did not show a significant effect on the computational cost of either method when compared to map size. Larger reading sizes were not considered as they are unlikely in real world applications.

\section{Discussion}
\label{sec:discussion}
PathSpace establishes the first method in literature to extract spline based probabilistic maps online and directly from noisy sensor data. The immediate strength lies in its ability to create a representational budget to optimize the required map size to the complexity of the environment. B-splines do however possess other properties that can further benefit an autonomous agent and can be the basis of a wide range of sub-systems that can make use of such innate representations to further benefit in terms of efficiency. 
While we have focussed on demonstrating the general feasibility and correctness of this approach, there is significant potential to encode more complex representations, such as using the spline as a path with a width rather than representing both bounds or otherwise detecting some other property of interest
that would not usually be captured from landmark detections, such as conditions or risk of the traversable area. 
The work presented here demonstrates the feasibility of this type of representation, and opens research in ways to further optimize and exploit this approach to mapping, 
providing both a more compact representation with richer semantic embeddings while more closely approximating the ground truth of the environment through interpolation.

\section{Conclusion}
\label{sec:conclusion}

In this paper, we introduce PathSpace, a novel framework to represent probabilistic structured environments in a continuous manner using B-splines in the context of SLAM with a focus on scalability and compactness.
We achieve a 16\% increase in mapping accuracy while reducing the required state size by 54\% with systems to optimize the map under this representation. PathSpace uses this system 
to limit the computational resources used, as well as interpolating beyond what the agent is able to sense to mitigate the effect of blind spots. Future research directions can further leverage the results of this paper by building on top of live systems, and exploit the inherit semantic elements of this methodology to develop smarter and more efficient control systems.

We demonstrate a potential implementation and benefits of such a system in the context of autonomous racing, showing how it can leverage the known structure of the 
track to generate a compact representation of the environment while surpassing the limitations of discrete point representations.

The potential reduction in computational cost offered by PathSpace's efficient representations paves the way for a new breed of SLAM algorithms based upon encoding critical and semantic environmental features. 
This would enable more scalable systems able to operate in larger areas seamlessly while offloading post-processing on other subsystems, allowing for more efficient systems both computationally and performance-wise.

\printbibliography

\end{document}